\documentclass[11pt]{article}
\usepackage{acl2016}
\usepackage{times}
\usepackage{latexsym}
\usepackage{booktabs} 
\usepackage{graphicx}
\usepackage{caption} 
\usepackage{subcaption}
\usepackage{CJKutf8}

\aclfinalcopy 
\def\aclpaperid{233} 

\title{Neural Name Translation Improves Neural Machine Translation}

\author{Xiaoqing Li, Jiajun Zhang and Chengqing Zong\\
	    National Laboratory of Pattern Recognition, Institute of Automation\\
	    Chinese Academy of Sciences\\
	    {\tt{xqli,jjzhang,cqzong}@nlpr.ia.ac.cn}
	    }

\date{}

\begin{document} \begin{CJK*}{UTF8}{gbsn}

\maketitle

\begin{abstract}
In order to control computational complexity, neural machine
translation (NMT)
systems convert all rare words outside the vocabulary into a single \textit{unk}
symbol. Previous solution \cite{luong-acl-15} resorts to use
multiple numbered \textit{unk}s to learn the correspondence between source and target rare
words. However, testing words unseen in the training corpus cannot be handled by
this method. And it also suffers from the noisy word alignment. In this
paper, we focus on a major type of rare words -- named entity (NE), and propose
to translate them with character level sequence to sequence model. The NE
translation model is further used to derive high quality NE alignment in the bilingual
training corpus. With the integration of NE translation and alignment modules, our NMT system is able to surpass the baseline system by 2.9 BLEU points on the Chinese to English task.

\end{abstract}

\section{Introduction}

Neural machine translation is a recently proposed approach to MT and has shown
competing results to conventional translation methods \cite{blunsom-emnlp-13,cho-emnlp-14,sutskever-nips-14}. Despite
several advantages over conventional methods, such as no domain knowledge
requirement, better generalization to novel translations and less memory
consumption, it has a significant weakness in handling rare words. In order
to control computational complexity, NMT systems convert all rare words
outside the vocabulary into a single \textit{unk} symbol. Such conversion makes them
unable to translate rare words. And those meaningless \textit{unk}s also increase the difficulty for the NMT model to learn
the correspondence between source and target words.

\begin{figure}[t] \centering
    \includegraphics[width=0.45\textwidth]{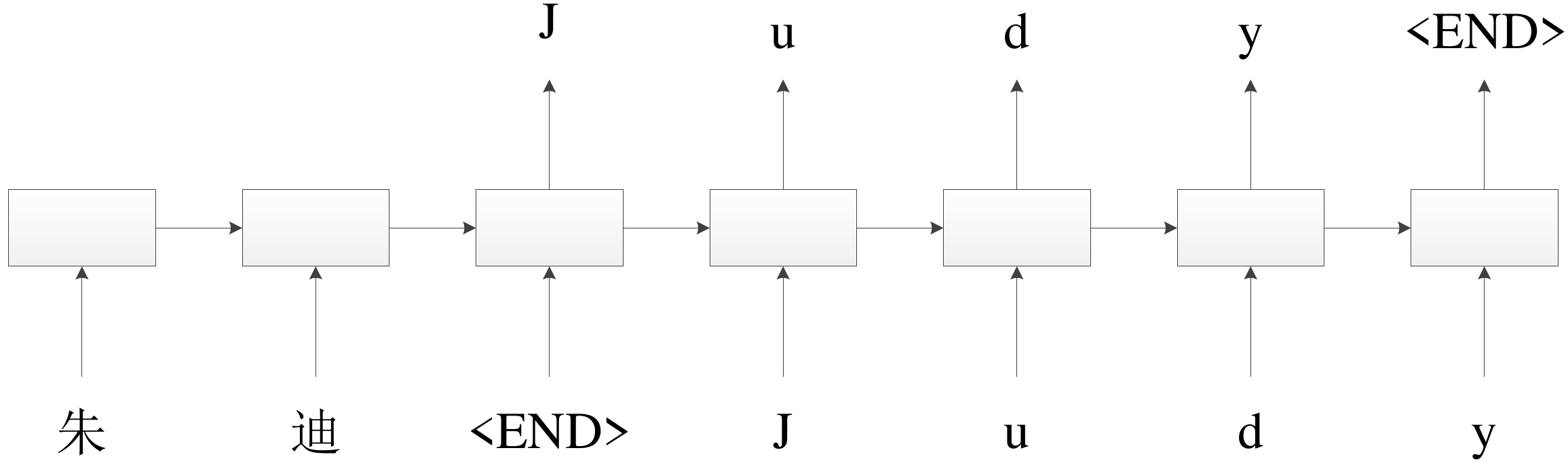} \caption{Character level
sequence to sequence model for NE translation. }
\end{figure}

To tackle this problem, Luong et al. \shortcite{luong-acl-15} propose to augment the \textit{unk} symbol
with alignment information. Their method allow the NMT system to learn, for
each \textit{unk} in the target sentence, the position of its corresponding word in the
source sentence. A post-processing step is adopted to translate target \textit{unk}s
with a dictionary.

This approach has been shown effective to handle rare words, but there are still some drawbacks. First, it cannot
translate words outside the dictionary. Second, it relies on noisy word alignment.
As known to all, automatic word alignment for rare words is far from perfect.
Wrong alignment will reduce the quality of the bilingual corpus to train NMT
model, and the dictionary extracted according to word alignment will also be
noisy.
Third, the content of rare words is totally ignored. Taking the following sentence as an example,

\begin{quote} 

Judy chases the pat with \textit{unk} .

\end{quote}

the translation of this sentence into Chinese will be quite different depending on whether the last word is a person name or a modifier describing some feature of the pat.

To solve the above problems, we propose to translate rare words with character
level sequence to sequence model, as shown in Figure 1. Due to the limitation
of existing resources, we limit our research in this paper to a major type of
rare words -- named entities.
With an NE translator trained on external NE list, we can derive high quality
NE alignment in the bilingual corpus, from which more NE pairs can be extract to further enhance the NE translator. 
Similar to Luong et al. \shortcite{luong-acl-15}, the aligned NE pairs are then replaced with
their type symbols and an NMT model is trained on the new data. A
post-processing step is employed to recover the translation of the replaced
NEs.

Our experiments demonstrate the effectiveness of this approach.  Evaluation on the Chinese to English translation task shows that
our integrated system surpasses the baseline system by 2.9 BLEU points, and brings an  improvement of 1.6 BLEU points over Luong's method.

\begin{figure*}[t] \centering \includegraphics[width=0.9\textwidth]{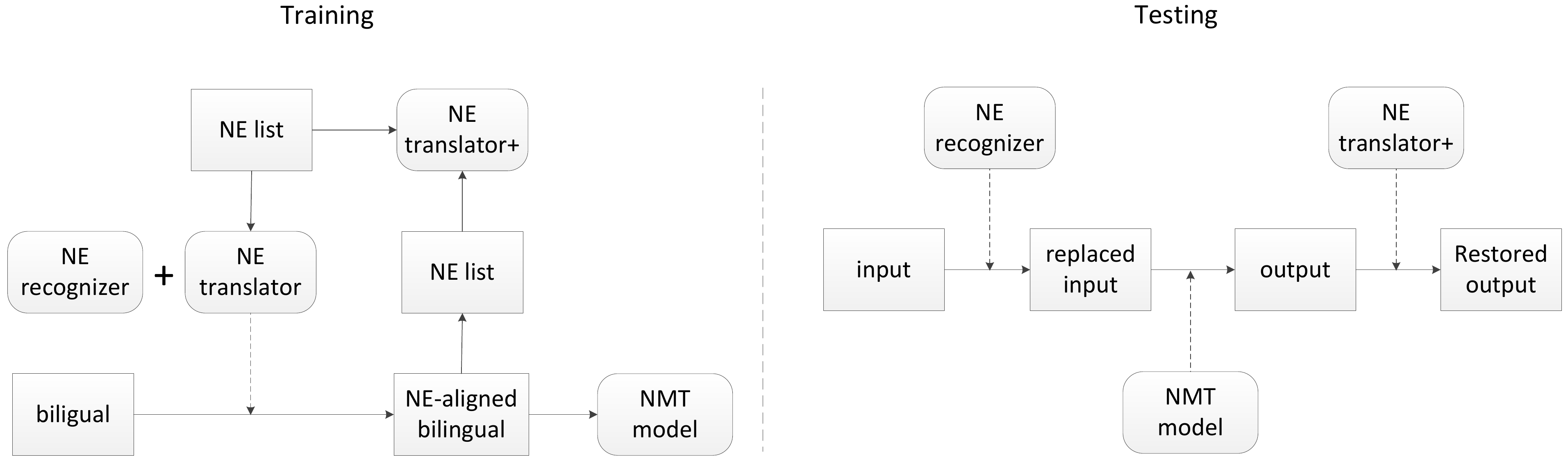}
\caption{System architecture to incorporate NE translation into neural MT} \end{figure*}

\section{Named Entity Translation and Alignment Model}

Figure 2 gives an overview of the architecture of our system. In the training
phase, we first train 
a neural NE translation system with a character-level sequence to sequence
model. The initial training data consists of NE translation pairs, which can be
obtained easily for many language pairs from the web. For example, we can
extract linked Wikipedia titles and filter them according to category
information. Then this NE translator together with a NE recognizer
are used to find aligned NE pairs in the bilingual corpus. A list of
NE pairs can be extracted from this corpus and it is combined with the original list
to build a stronger NE translator. The aligned NE pairs in the bilingual corpus will be replaced  with
corresponding type symbols, resulting in sentence pairs like the following
example,

\begin{quote} 

ZH: LOC1 重新\ 开放\ 驻\ LOC2 大使馆

EN: LOC1 reopens embassy in LOC2

\end{quote}

Finally, this new data after replacement will be used to train an NMT
model.

In the testing phase, the NEs in the input sentence are first recognized and
replaced with type symbols. After translated by the NMT model, the NE symbols
in the output will be replaced with the translation of original NEs, which is
generated by the NE translation module.

\subsection{Named Entity Translator}

The model we adopt to translate NE is a character level sequence to sequence model. It
maps a source NE $s = (s_1,s_2,...,s_m)$ into a target NE
$t=(t_1,t_2,...,t_n)$ with a single neural network as follows,

\[ p(t|s) = \prod_{i=1}^n p(t_i|t_{<i},s) \]

where the conditional probability is paramiterized with the encoder decoder
framework. The encoder reads the source character sequence and encodes it into
a sequence of hidden states. Then the decoder generates the target NE character by character based on the target hidden states. In this paper, we adopt the implementation of Bahdanau et al. \shortcite{bahdanau-iclr-15} which introduces an
attention mechanism while predicting each unit in target sequence. Please look
into it for more detail.

As pointed out by previous work, the total computational
complexity will grow almost proportional to the target vocabulary in the sequence to sequence model.
Fortunately, in the scenario of character level NE translation, the 
vocabulary size is only hundreds (such as English) to thousands (such as
Chinese). While in the case of word level NMT, the vocabulary size is often
hundreds of thousands to millions. So NMT systems usually limit vocab size to
tens of thousands to make computation feasible. In character level model,
there is no such need.

\subsection{Named Entity Alignment}

Named entity alignment based on bilingual corpus alone is a hard task. Since
a lot of NEs appear only a few times in the corpus, we cannot collect enough
statistical evidence to infer reliable alignments with traditional word
alignment model. Previous work \cite{huang-multiner-03,feng-emnlp-04} design multiple features, including
translation score, transliteration score, distortion score etc., and use
iterative training to discover aligned NEs in bilingual corpus. 

However, if we have a high quality and high coverage NE list, together with a
powerful NE translation model, things could be much easier. We can learn a NE translator from the list, then use it to translate
recognized NEs in one language and compare it with word sequence (up to trigram) in the other
language. The longest common subsequence between NE translation candidate and
target word sequence is adopted for similarity matching in this
paper,

\[
LCS(c,t)=\frac{1}{2}(|c|+|t|-ED(c,t))
\]
\[
sim(c,t) = LCS(c,t)/|c|
\]

where $ED$ is the edit distance and $|x|$ is the length of $x$. For example, the
longest common subsequence between 'bolin' and 'berlin' is 'blin'. 

The final NE　alignment result is the union of bi-directional matching, i.e.
matching source NE with target word sequence and vice versa. We do not 
match bilingual NEs directly because the automatic NE recognition is not good enough
and some NEs are not recognized in each language.

It has to mention that we don't have a list of numerical and temporal expressions to train a corresponding
translation model before alignment. To calculate the similarity score, we
carry the following conversion for both Chinese and English numerical
expression,

\begin{itemize}

\item[-] Convert Chinese and English numbers one to nine to abric numbers, discard
  all other characters. For example, '百分之四点二' (4.2\%) will be converted into 42,
and 4,200 will also be converted into 42.

\item[-] Add a few rules to handle exceptions such as month.

\end{itemize}

The above conversion is used for the sake of alignment. After NE pairs are extracted according to the alignment, they are used to
train the NE translation model to handle NEs in testing data.

\section{Experiments}

We evaluate our method on the Chinese to English translation task.
Translation quality is measured by the BLEU metric \cite{papineni-acl-02}.

\subsection{Settings}

The bilingual data to train the NMT model is selected from LDC, which
contains about 0.6M sentence pairs. To avoid spending too much training
time on long sentences, all sentence pairs longer than 50 words either
on the source side or on the target side are discarded. The initial NE pairs
are extracted from the Wikipedia, which consist about 350k entries. We use an in-house developed NE recognizer for Chinese and Stanford NER \cite{finkel-acl-05} for English.

The NIST 03 dataset is chosen as the development set, which is used to
monitoring the training process and decide the early stop condition. And
the NIST 04 to 06 are used as our testing set.

\subsection{Training Details}

The hyperparameters used in our network are described as
follows. We limit both the source and target vocabulary to 30k in our
experiments. Names inside the vocabulary are not handled. The number of hidden units is 1,000 for both the encoder
and decoder. And the embedding dimension is 500 for all source and
target tokens. The network parameters are updated with the adadelta
algorithm and the learning rate is set to $10^{-4}$. The above setting is used 
both in the character level NE translation and word level sentence translation.

\begin{table}[] \centering \begin{tabular}{llll} \toprule  & N/T & PER & LOC \\ 
\midrule 
translation & 0.71  & 0.28 & 0.35 \\ 
trans. + lex. table & 0.78 & 0.48 & 0.70 \\ 
\midrule 
     alignment & 0.97 & 0.93 & 0.96 \\
      \bottomrule
      \end{tabular} 
      \caption{Translation and alignment performance with neural NE translation model } 
      \end{table}

\subsection{Name Translation and Alignment Performance}

\begin{table*}[] 
\centering \begin{tabular}{llllll}
\toprule 
System & 03 (dev) & 04 & 05 & 06 & Average\\ 
\midrule 
baseline & 25.65   & 28.94 & 25.13 & 27.86 & 26.90 \\
unk rep. & 27.63 & 30.02 & 26.42 & 28.72 & 28.20 \\ 
NE rep. & 27.90 & 30.67 & 28.20 & 29.42 & 29.05 \\ 
unk+NE rep. & \textbf{29.01} & \textbf{31.33} & \textbf{28.80} & \textbf{30.08} & \textbf{29.80} \\
\bottomrule
\end{tabular} 
\caption{Translation results for different systems}

\end{table*}

Because the recognition performance and cross-lingual consistency are not good
for organization names, we ignore this type and only handle numerical/temporal
expressions, person names and location names in this paper. To evaluate the NE
translation performance, we randomly extract 100 instances from the NIST
testing data for each type, and manually find their translations in the
reference. Whereas the NE alignment performance is evaluated on the same amount of samples extracted from the training data. The results are shown in Table 1.

It could be seen from the table that the translation accuracy is relatively
low. Since accuracy is calculated in word level, the NE translation is
regarded wrong even when there is only one letter different. In order to
improve the NE translation accuracy in the post-processing step, we propose to use the lexical table
extracted from the bilingual data if the NE could be found, otherwise the
neural NE translation model will be used. On the other hand, the alignment
accuracy is quite high. And most alignment errors relate to wrong word
segmentation or NE recognition according to our investigation.

\subsection{Sentence Translation Performance}

We compare the translation performance of our method with that of Luong et al.
\shortcite{luong-acl-15}. The results are shown in Table 2. The baseline system we
adopt is the  attention-based model proposed in Bahdanau et al. \shortcite{bahdanau-iclr-15} . It can be seen that only replacing rare NEs with our
method results in a better performance than replacing all rare words with
Luong's method. After combining the two methods, i.e., replacing NEs with our method and replacing other rare words with Luong's method, we could obtain an extra performance boost of 0.75 BLEU points, and the final performance surpass the baseline by 2.9 BLEU points on average. It has to be mentioned that Luong's method is not as effective on the Chinese-English language pair as reported on the French-English language pair. A possible reason is the automatic word alignment quality is worse on the former language pair.

\section{Related Work}

Inability to handle rare words is a significant defect of NMT systems. And it has attracted much attention recently. Besides the work of Luong \shortcite{luong-acl-15}, Jean et al. \shortcite{jean-acl-15} propose  to  directly  use  large  vocabulary  with  a  method  based on importance sampling.  As pointed out in their paper, their method is complementary and can be used together with replacement methods. Sennrich et al. \shortcite{sennrich-arxiv-15} propose to represent rare words as sequences
of subword units, and compares different techniques to segment words into
subwords. Wang et al. \shortcite{wang-arxiv-15} use a hierarchical structure
for NMT, in which word representations are derived from character
representations.

The problem of NE translation has been studied for a long time. Knight and
Graehl \shortcite{knight-cl-98} study it with probabilistic finite-state transducers. Li et al. \shortcite{li-acl-04}
present a joint source-channel model for direct orthographical mapping
without intermediate phonemic representation. Freitag and Khadivi
\shortcite{freitag-emnlp-07} propose a technique which combines conventional
MT methods with a single layer perceptron. Deselaers et al.
\shortcite{deselaers-wssmt-09} use deep belief networks for machine
transliteration. There are also some previous works trying to integrate NE
translation into traditional MT systems. Hermjakob et al.
\shortcite{hermjakob--acl-08} study the problem of "when to transliterate".
Li et al. \shortcite{li-acl-13} propose to combine two copies of training
data, the original one and the one with aligned NEs replaced. 

\section{Conclusion}

In this paper, we enhance the ability of NMT system to handle rare words by
incorporating NE translation and alignment modules.  With the help of extra NE
list and NE recognizer, our method is able to produce high quality NE
alignments, and thus improves the data quality to train NE translation and
sentence translation model. Experimental results show that our approach can significantly improve the translation performance.

\bibliography{acl2016}

\begin{thebibliography}{}

\bibitem[\protect\citename{Bahdanau \bgroup et al.\egroup
  }2015]{bahdanau-iclr-15}
Dzmitry Bahdanau, Kyunghyun Cho, and Yoshua Bengio.
\newblock 2015.
\newblock Neural machine translation by jointly learning to align and
  translate.
\newblock In {\em ICLR 2015}.

\bibitem[\protect\citename{Cho \bgroup et al.\egroup }2014]{cho-emnlp-14}
Kyunghyun Cho, Bart van Merrienboer, Caglar Gulcehre, Dzmitry Bahdanau, Fethi
  Bougares, Holger Schwenk, and Yoshua Bengio.
\newblock 2014.
\newblock Learning phrase representations using rnn encoder--decoder for
  statistical machine translation.
\newblock In {\em Proceedings of the 2014 Conference on Empirical Methods in
  Natural Language Processing (EMNLP)}, pages 1724--1734, Doha, Qatar, October.
  Association for Computational Linguistics.

\bibitem[\protect\citename{Deselaers \bgroup et al.\egroup
  }2009]{deselaers-wssmt-09}
Thomas Deselaers, Sa{\v{s}}a Hasan, Oliver Bender, and Hermann Ney.
\newblock 2009.
\newblock A deep learning approach to machine transliteration.
\newblock In {\em Proceedings of the Fourth Workshop on Statistical Machine
  Translation}, pages 233--241. Association for Computational Linguistics.

\bibitem[\protect\citename{Feng \bgroup et al.\egroup }2004]{feng-emnlp-04}
Donghui Feng, Yajuan Lv, and Ming Zhou.
\newblock 2004.
\newblock A new approach for english-chinese named entity alignment.
\newblock In Dekang Lin and Dekai Wu, editors, {\em Proceedings of EMNLP 2004},
  pages 372--379, Barcelona, Spain, July. Association for Computational
  Linguistics.

\bibitem[\protect\citename{Finkel \bgroup et al.\egroup }2005]{finkel-acl-05}
Jenny~Rose Finkel, Trond Grenager, and Christopher Manning.
\newblock 2005.
\newblock Incorporating non-local information into information extraction
  systems by gibbs sampling.
\newblock In {\em Proceedings of the 43rd Annual Meeting of the Association for
  Computational Linguistics (ACL'05)}, pages 363--370, Ann Arbor, Michigan,
  June. Association for Computational Linguistics.

\bibitem[\protect\citename{Freitag and Khadivi}2007]{freitag-emnlp-07}
Dayne Freitag and Shahram Khadivi.
\newblock 2007.
\newblock A sequence alignment model based on the averaged perceptron.
\newblock In {\em Proceedings of the 2007 Joint Conference on Empirical Methods
  in Natural Language Processing and Computational Natural Language Learning
  (EMNLP-CoNLL)}, pages 238--247, Prague, Czech Republic, June. Association for
  Computational Linguistics.

\bibitem[\protect\citename{Hermjakob \bgroup et al.\egroup
  }2008]{hermjakob--acl-08}
Ulf Hermjakob, Kevin Knight, and Hal Daum\'{e}~III.
\newblock 2008.
\newblock Name translation in statistical machine translation - learning when
  to transliterate.
\newblock In {\em Proceedings of ACL-08: HLT}, pages 389--397, Columbus, Ohio,
  June. Association for Computational Linguistics.

\bibitem[\protect\citename{Huang \bgroup et al.\egroup
  }2003]{huang-multiner-03}
Fei Huang, Stephan Vogel, and Alex Waibel.
\newblock 2003.
\newblock Automatic extraction of named entity translingual equivalence based
  on multi-feature cost minimization.
\newblock In {\em Proceedings of the ACL 2003 Workshop on Multilingual and
  Mixed-language Named Entity Recognition}, pages 9--16, Sapporo, Japan, July.
  Association for Computational Linguistics.

\bibitem[\protect\citename{Jean \bgroup et al.\egroup }2015]{jean-acl-15}
S\'{e}bastien Jean, Kyunghyun Cho, Roland Memisevic, and Yoshua Bengio.
\newblock 2015.
\newblock On using very large target vocabulary for neural machine translation.
\newblock In {\em Proceedings of the 53rd Annual Meeting of the Association for
  Computational Linguistics and the 7th International Joint Conference on
  Natural Language Processing (Volume 1: Long Papers)}, pages 1--10, Beijing,
  China, July. Association for Computational Linguistics.

\bibitem[\protect\citename{Kalchbrenner and Blunsom}2013]{blunsom-emnlp-13}
Nal Kalchbrenner and Phil Blunsom.
\newblock 2013.
\newblock Recurrent continuous translation models.
\newblock In {\em Proceedings of the 2013 Conference on Empirical Methods in
  Natural Language Processing}, pages 1700--1709, Seattle, Washington, USA,
  October. Association for Computational Linguistics.

\bibitem[\protect\citename{Knight and Graehl}1998]{knight-cl-98}
Kevin Knight and Jonathan Graehl.
\newblock 1998.
\newblock Machine transliteration.
\newblock {\em Computational Linguistics}, 24(4):599--612.

\bibitem[\protect\citename{Li \bgroup et al.\egroup }2004]{li-acl-04}
Haizhou Li, Min Zhang, and Jian Su.
\newblock 2004.
\newblock A joint source-channel model for machine transliteration.
\newblock In {\em Proceedings of the 42nd Meeting of the Association for
  Computational Linguistics (ACL'04), Main Volume}, pages 159--166, Barcelona,
  Spain, July.

\bibitem[\protect\citename{Li \bgroup et al.\egroup }2013]{li-acl-13}
Haibo Li, Jing Zheng, Heng Ji, Qi~Li, and Wen Wang.
\newblock 2013.
\newblock Name-aware machine translation.
\newblock In {\em Proceedings of the 51st Annual Meeting of the Association for
  Computational Linguistics (Volume 1: Long Papers)}, pages 604--614, Sofia,
  Bulgaria, August. Association for Computational Linguistics.

\bibitem[\protect\citename{Ling \bgroup et al.\egroup }2015]{wang-arxiv-15}
Wang Ling, Isabel Trancoso, Chris Dyer, and Alan~W Black.
\newblock 2015.
\newblock Character-based neural machine translation.
\newblock {\em arXiv preprint arXiv:1511.04586}.

\bibitem[\protect\citename{Luong \bgroup et al.\egroup }2015]{luong-acl-15}
Thang Luong, Ilya Sutskever, Quoc Le, Oriol Vinyals, and Wojciech Zaremba.
\newblock 2015.
\newblock Addressing the rare word problem in neural machine translation.
\newblock In {\em Proceedings of the 53rd Annual Meeting of the Association for
  Computational Linguistics and the 7th International Joint Conference on
  Natural Language Processing (Volume 1: Long Papers)}, pages 11--19, Beijing,
  China, July. Association for Computational Linguistics.

\bibitem[\protect\citename{Papineni \bgroup et al.\egroup
  }2002]{papineni-acl-02}
Kishore Papineni, Salim Roukos, Todd Ward, and Wei-Jing Zhu.
\newblock 2002.
\newblock Bleu: a method for automatic evaluation of machine translation.
\newblock In {\em Proceedings of 40th Annual Meeting of the Association for
  Computational Linguistics}, pages 311--318, Philadelphia, Pennsylvania, USA,
  July. Association for Computational Linguistics.

\bibitem[\protect\citename{Sennrich \bgroup et al.\egroup
  }2015]{sennrich-arxiv-15}
Rico Sennrich, Barry Haddow, and Alexandra Birch.
\newblock 2015.
\newblock Neural machine translation of rare words with subword units.
\newblock {\em arXiv preprint arXiv:1508.07909}.

\bibitem[\protect\citename{Sutskever \bgroup et al.\egroup
  }2014]{sutskever-nips-14}
Ilya Sutskever, Oriol Vinyals, and Quoc~VV Le.
\newblock 2014.
\newblock Sequence to sequence learning with neural networks.
\newblock In {\em Advances in neural information processing systems}, pages
  3104--3112.

\end{thebibliography}
\bibliographystyle{acl2016} 

\end{CJK*}
\end{document}